\documentclass[11pt]{article}
\usepackage[preprint]{acl}
\usepackage{times,latexsym}
\usepackage[T1]{fontenc}
\usepackage[utf8]{inputenc}
\usepackage{microtype,inconsolata}
\usepackage{amsmath,amssymb,graphicx,booktabs,multirow}
\usepackage{listings,placeins,caption,tabularx}
\usepackage{xurl}
\title{RIPPLE: Recovery Internalization through Process-Level Policy Learning and Effect-Aware Credit for Long-Horizon Agentic Coding}
\author{
\textbf{Jingxuan Xu}$^{1,\star}$,
\textbf{Zihang Xu}$^{2,\star}$,
\textbf{Yuechen Luo}$^{3,\star}$,
\textbf{Ziqi Miao}$^{2}$,
\textbf{Wenqiang Zhu}$^{4}$ \\
\textbf{Weihao Li}$^{4}$,
\textbf{Junhao Fan}$^{3}$,
\textbf{Xinshun Feng}$^{2}$,
\textbf{Zhiyi Lai}$^{4}$,
\textbf{Yanan Wu}$^{4}$ \\
\textbf{Haoyang Huang}$^{4}$,
\textbf{Ken Deng}$^{4,\dagger}$ \\
\normalfont\large $^1$ National University of Singapore, Singapore \\
\normalfont\large $^2$ Shanghai Artificial Intelligence Laboratory, Shanghai, China \\
\normalfont\large $^3$ Tsinghua University, Beijing, China \\
\normalfont\large $^4$ Independent Researchers \\
\normalfont\normalsize \texttt{e1553956@u.nus.edu}
}
\hypersetup{
  pdftitle={RIPPLE: Recovery Internalization through Process-Level Policy Learning and Effect-Aware Credit for Long-Horizon Agentic Coding},
  pdfauthor={Jingxuan Xu, Zihang Xu, Yuechen Luo, Ziqi Miao, Wenqiang Zhu, Weihao Li, Junhao Fan, Xinshun Feng, Zhiyi Lai, Yanan Wu, Haoyang Huang, Ken Deng}
}
\begin{document}
\maketitle
\begingroup
\renewcommand{\thefootnote}{}
\makeatletter
\renewcommand{\@makefntext}[1]{\noindent #1}
\makeatother
\footnotetext{$^{\star}$ Equal contribution.\\$^{\dagger}$ Corresponding author.}
\endgroup
\begin{abstract}
Long-horizon coding agents can observe a failure and generate a plausible critique yet continue along the same incorrect trajectory. We call this the Critique-to-Action Handoff Gap. We introduce \textsc{RIPPLE}, a recovery-oriented training framework that internalizes process-level guidance into a single self-recovering policy. A prefix-only teacher provides evidence-grounded recovery supervision without access to future actions or terminal outcomes. After lightweight distillation, reinforcement learning assigns stage-specific credit to reflection triggering, diagnosis, and corrective action. Local validity signals gate trajectory advantages so that intermediate recovery decisions do not indiscriminately inherit terminal outcomes. Across four repository-level software-engineering benchmarks, \textsc{RIPPLE} improves the base policy by an arithmetic mean of 4.8 score points, with no auxiliary critic at deployment. Controlled ablations isolate the benefits of local, stage-aligned credit, and matched-prefix interventions show that critique semantics affect subsequent correction. Recovery gains concentrate where the prefix contains sufficient diagnostic evidence and the repository state remains reversible within the remaining budget.
\end{abstract}
\section{Introduction}
\begin{figure}[t]
\centering
\includegraphics[width=\columnwidth]{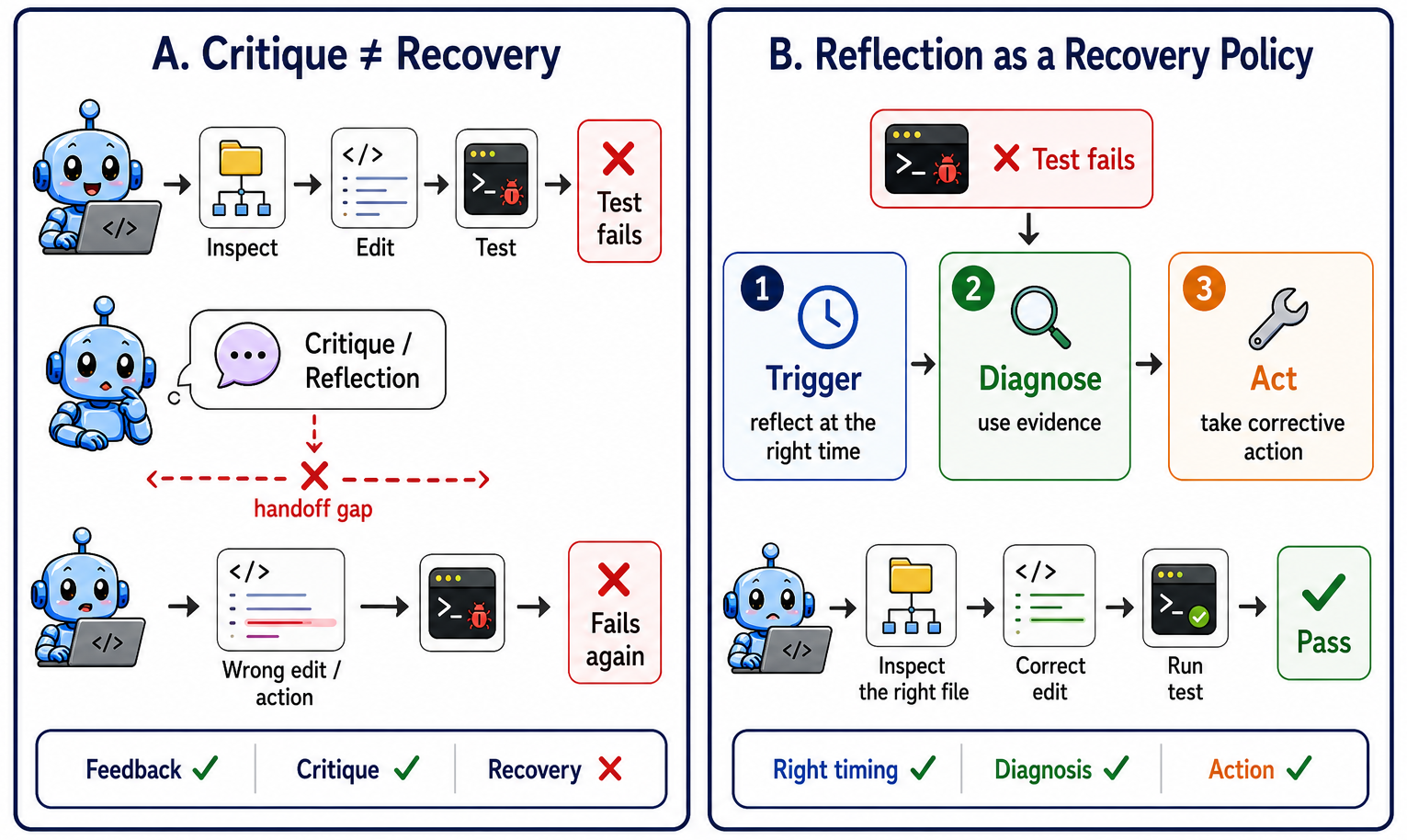}
\caption{\textbf{The critique-to-action handoff gap.} Failure feedback may elicit an evidence-grounded critique without changing subsequent behavior (left). \textsc{RIPPLE} learns a recovery policy over reflection triggering, diagnosis, and corrective action (right).}
\label{fig:motivation}
\end{figure}
Large language model-based coding agents now explore repositories, edit files, execute tests, and revise their behavior using environmental feedback. Repository-level benchmarks and agent frameworks enable these capabilities over long trajectories in realistic software environments \citep{jimenez2024swebench,yang2024sweagent,pan2024swegym,wei2025swerl,golubev2025longcontext,ding2026swereplay}. However, observing feedback does not necessarily imply using it effectively. When a test invalidates an earlier assumption, an agent may continue editing in the same incorrect direction, repeatedly inspect irrelevant files, or fail to revise its plan \citep{zhao2026failureprocessanatomycli,li2026codetracer}. Although modern agents can close the action--observation loop, reliable recovery from trajectory deviations remains challenging.

Reflection and critique provide a natural mechanism for addressing these failures. Self-generated feedback, verbal memory, tool-assisted critique, and execution-based debugging can improve model outputs \citep{madaan2023selfrefine,shinn2023reflexion,gou2024critic,chen2023selfdebug}. Recent approaches further train self-correction through supervised learning and reinforcement learning (RL) \citep{kumar2024score,yuan2025agentr,jin2025reveal,jiang2026reflexicoder,lin2026icrl}. However, improved task performance does not establish that a critique guides the next action \citep{shinn2023reflexion,zhao2026failureprocessanatomycli}. As Figure~\ref{fig:motivation} illustrates, an agent may correctly identify a failure yet take a follow-up action inconsistent with its diagnosis. We refer to this discrepancy as the \emph{Critique-to-Action Handoff Gap}: an evidence-grounded critique remains behaviorally inert instead of inducing an effective corrective change.

This gap exposes a mismatch between the functional structure of recovery and existing training objectives. Critique imitation supervises what the model says about a failure, without establishing whether subsequent actions implement its diagnosis. Terminal rewards supervise trajectory success, but cannot isolate whether reflection was triggered appropriately, the diagnosis was correct, or the corrective action was effective. Process supervision and credit-assignment methods provide finer-grained signals \citep{lightman2023verify,ekbote2025murphy,guo2026pivoarl}, but generally assess intermediate reasoning or redistribute trajectory outcomes rather than separately optimize the functional stages of recovery. Addressing the handoff gap requires jointly learning when to initiate recovery, which diagnosis the evidence supports, and how it should alter subsequent behavior.

We propose \textsc{RIPPLE}, a recovery-oriented training framework for long-horizon coding agents. A trajectory-prefix-conditioned teacher supplies recovery boundaries and reference critiques using only information available at each decision point. Lightweight distillation initializes task execution and interleaved reflection. Recovery-oriented RL then decomposes recovery into reflection triggering, evidence-grounded diagnosis, and corrective action, assigning a distinct process signal to each stage. An executable verifier supervises task correctness, while local validity gates determine whether each recovery span inherits positive or negative trajectory advantage. This design targets the functional contribution of intermediate decisions rather than their co-occurrence with successful trajectories. All auxiliary supervision is confined to training, leaving a single self-recovering policy at inference.

Experiments show an arithmetic mean improvement of 4.8 points across four software-engineering benchmarks without an online critic. Ablations and matched-prefix interventions assess the functional contribution of recovery. Our contributions are:
\begin{itemize}
\item \textbf{Critique-to-action recovery formulation.} We identify the handoff gap and formulate recovery as an interleaved policy that coordinates when to reflect, what to diagnose, and how to correct subsequent behavior.
\item \textbf{Stage-specific recovery optimization.} We combine prefix-only supervision with effect-aware credit for triggering, diagnosis, and corrective action while retaining terminal supervision over task completion.
\item \textbf{Mechanism-grounded empirical evaluation.} End-to-end comparisons, credit-assignment ablations, and matched-prefix interventions assess issue resolution and the functional contribution of critiques, including the evidence and repository-state conditions under which recovery remains effective.
\end{itemize}

\section{Related Work}

\paragraph{Software engineering agents.}
Repository-level benchmarks establish real-world issue resolution and
executable implementation tasks as important settings for evaluating
language-model agents \citep{jimenez2024swebench,yu2026swenexus,yang2025codeintelligence}.
Complementary evaluations cover repository-level completion
\citep{yang2024execrepobench,deng2024r2c2,liu2025m2rc}, web coding
\citep{lei2026webcompass}, multi-turn tool use \citep{wang2024mtubench},
and trajectory-level engineering constraints \citep{zhu2026anchor}.
Agent capabilities improve through specialized interaction interfaces,
executable training environments, and reinforcement learning over
repository-level trajectories
\citep{yang2024sweagent,pan2024swegym,wei2025swerl}, with multi-stage
curricula combining supervised and reinforcement fine-tuning
\citep{zhan2025katcoder}.
Reusing intermediate agent trajectories improves inference efficiency by
preserving progress across attempts \citep{ding2026swereplay}, while trajectory analyses
assess reasoning--action mismatches and localize failure onset in
long-horizon executions \citep{bouzenia2025trajectories,li2026codetracer}.
\textsc{RIPPLE} learns an internal recovery policy that responds when new
evidence invalidates the agent's current hypothesis and links diagnosis
to subsequent corrective behavior.

\paragraph{Reflection and critique internalization.}
Self-generated feedback, verbal memory, tool-assisted critique, and
execution results support iterative revision of model outputs across
successive attempts
\citep{madaan2023selfrefine,shinn2023reflexion,gou2024critic,chen2023selfdebug,li2026relook}.
Recent work evaluates code critiques \citep{zhang2025codecriticbench}
and trains models to self-correct or recover from earlier errors through
supervised learning and reinforcement learning
\citep{yuan2025agentr,jin2025reveal,lin2026icrl}.
OPRD internalizes reminder-conditioned guidance through on-policy
distillation \citep{luo2026oprd}.
External intervention systems detect problematic behavior in coding-agent
trajectories and provide targeted course correction during execution
\citep{nanda2026wink}.
\textsc{RIPPLE} addresses the \emph{Critique-to-Action Handoff Gap} by
internalizing training-time recovery guidance into a selective policy,
with separate credit for when to reflect, what to diagnose, and how to
correct subsequent behavior.

\paragraph{Process supervision and credit assignment.}
Process supervision provides intermediate learning signals beyond
terminal task outcomes \citep{lightman2023verify}, including generative
reward modeling \citep{zhang2024generative,chen2025rmr1,xu2026flare},
step-level trajectory calibration, and behavioral critics learned from
sparse outcomes \citep{wang2025steca,wang2026rubriccritic}.
Credit assignment refines supervision in multi-turn trajectories through
retrospective reward propagation, pivotal-state identification, and
turn-level value estimation
\citep{ekbote2025murphy,guo2026pivoarl,tao2026trace}, as well as signals
for functional roles and task dependencies \citep{miao2026seeing,li2026weave}.
\textsc{RIPPLE} retains terminal task rewards and gates their advantages
locally using stage-specific validity signals for reflection triggering,
evidence-grounded diagnosis, and corrective-action effectiveness. Each
signal applies to the corresponding recovery span, connecting local
credit to its behavioral role and reducing terminal reward broadcasting
and cross-stage credit mismatch.

\begin{figure*}[t]
    \centering
    \includegraphics[width=\textwidth]{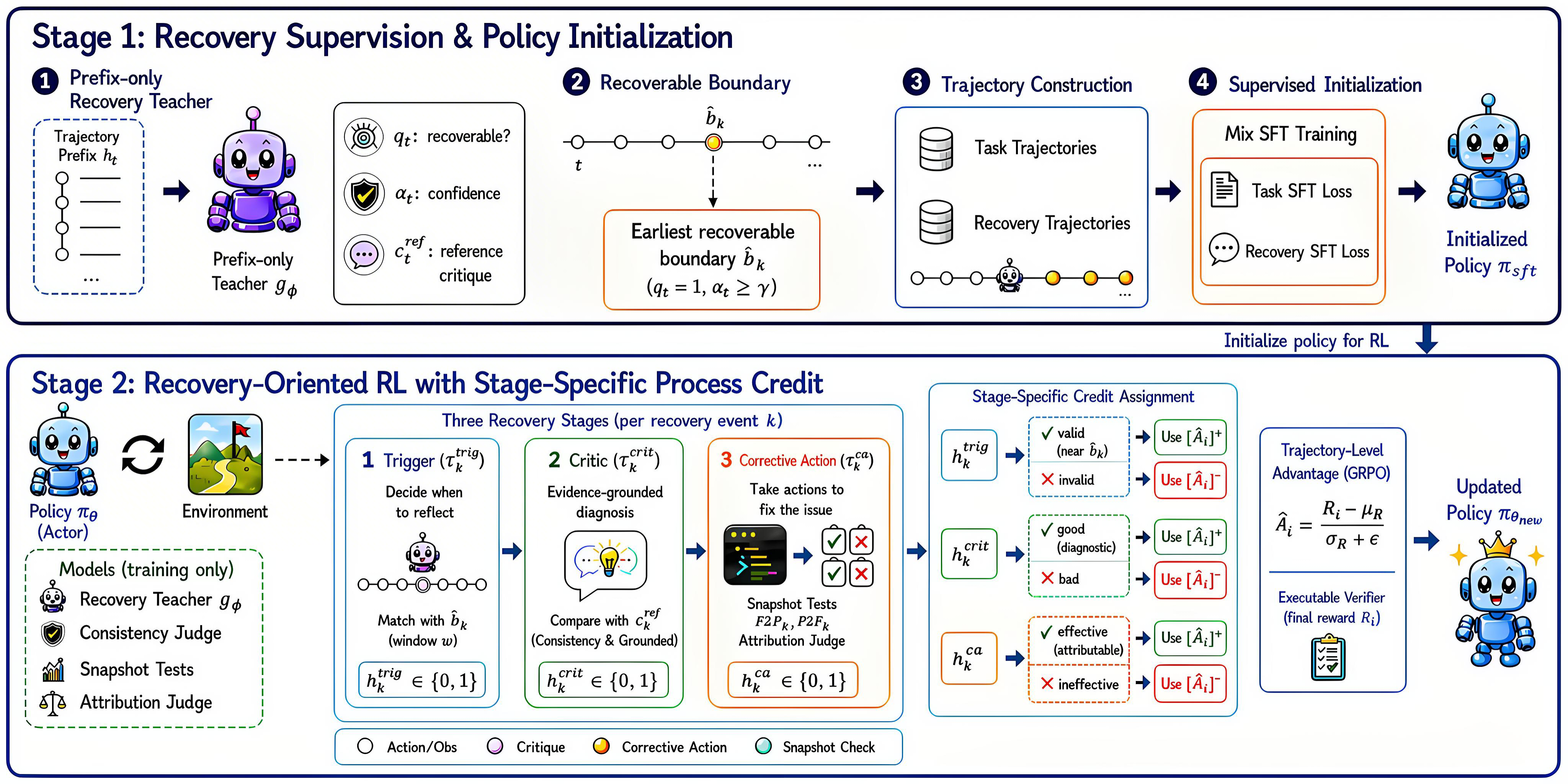}
    \caption{
    Overview of \textsc{RIPPLE}.
    A prefix-only recovery teacher supports policy initialization and process
    evaluation. The policy is then optimized through recovery-oriented
    reinforcement learning, which assigns stage-specific process credit to
    trigger, diagnosis, and corrective-action decisions, with the executable
    verifier as the sole source of task reward.
    }
    \label{fig:ripple_framework}
\end{figure*}

\section{\textsc{RIPPLE}}
\label{sec:method}

\subsection{Overall Framework}
\label{sec:overall}

\textsc{RIPPLE} aims to transform external process supervision into an
internal self-recovery policy for long-horizon coding agents. Given a
repository-level task \(x\), the agent interacts with an execution
environment and produces a trajectory
\(\tau=(x,a_1,o_1,\ldots,a_T,o_T)\), where \(a_t\) is an agent action and
\(o_t\) is the corresponding environmental observation. In addition to
ordinary task actions, the agent may insert explicit self-critiques into
its execution trajectory.

We decompose the \(k\)-th recovery episode into three functionally
distinct decision spans:
\begin{equation}
    e_k =
    \left(
        T_k^{\mathrm{trig}},
        T_k^{\mathrm{crit}},
        T_k^{\mathrm{ca}}
    \right),
    \label{eq:recovery_episode}
\end{equation}
where \(T_k^{\mathrm{trig}}\) contains the decision tokens that initiate
reflection, \(T_k^{\mathrm{crit}}\) contains the resulting diagnosis, and
\(T_k^{\mathrm{ca}}\) contains the subsequent corrective actions. These
stages respectively determine when recovery should begin, why the
trajectory has deviated, and how the agent should revise its behavior.

Figure~\ref{fig:ripple_framework} summarizes the two training stages: distillation initializes interleaved task execution and recovery, and RL refines the three recovery decisions. The executable verifier supplies scalar task reward; frozen process evaluators localize credit within each episode. Only the learned policy is used at inference.

\subsection{Recovery Supervision and Policy Initialization}
\label{sec:initialization}

\paragraph{Prefix-only recovery teacher.}
We train a lightweight generative model \(g_{\phi}\) as a training-time
recovery teacher. Given only the trajectory prefix \(h_t\), the teacher
produces
\begin{equation}
    g_{\phi}(h_t)
    =
    \left(
        q_t,\alpha_t,c_t^{\mathrm{ref}}
    \right),
    \label{eq:teacher_output}
\end{equation}
where \(q_t\in\{0,1\}\) indicates whether a recoverable deviation is
observable, \(\alpha_t\in[0,1]\) is the confidence of this judgment, and
\(c_t^{\mathrm{ref}}\) is an evidence-grounded reference critique. The
reference critique identifies the error type, relevant evidence,
invalidated assumption, and recovery direction.

Let \(\mathcal W_k\) denote the candidate region associated with the \(k\)-th recovery event. Its teacher-estimated recovery onset is
\begin{equation}
    \widehat b_k
    =
    \min
    \left\{
        t\in\mathcal W_k
        \;\middle|\;
        q_t=1,\;
        \alpha_t\geq\gamma
    \right\},
    \label{eq:recovery_boundary}
\end{equation}
where \(\gamma\) is a confidence threshold. When the set is non-empty, the
event-level reference critique is
\(c_k^{\mathrm{ref}}=c_{\widehat b_k}^{\mathrm{ref}}\); otherwise, the
teacher abstains.

Although complete trajectories are used offline to identify candidate
recovery events, the corresponding teacher outputs are regenerated from
their prefixes. The teacher prediction therefore does not condition on
future actions, the final patch, or the terminal verifier outcome.
Details appear in Appendix~\ref{app:supervision}.

\paragraph{Policy initialization.}
We construct an initial policy from task trajectories and recovery
trajectories. Task trajectories teach repository exploration, test
interpretation, code modification, and patch verification. Recovery
trajectories insert teacher-generated critiques at recoverable positions
and preserve the corresponding corrective continuations, providing an
initial prior over interleaved self-recovery.

Let \(\mathcal D_{\mathrm{task}}\) and
\(\mathcal D_{\mathrm{rec}}\) denote the two trajectory sets. We optimize
\begin{equation}
\begin{aligned}
    \mathcal L_{\mathrm{init}}
    =
    &\,
    \mathcal L_{\mathrm{SFT}}
    \left(
        \mathcal D_{\mathrm{task}}
    \right)
    \\
    &+
    \lambda_{\mathrm{rec}}
    \mathcal L_{\mathrm{SFT}}
    \left(
        \mathcal D_{\mathrm{rec}}
    \right),
\end{aligned}
\label{eq:init_objective}
\end{equation}
yielding the initialized policy \(\pi_{\mathrm{sft}}\). This stage teaches
the basic form of task execution and recovery behavior, but does not
determine whether a generated critique is timely, correct, or
behaviorally effective. These properties are optimized through
reinforcement learning.

\subsection{Recovery-Oriented Reinforcement Learning}
\label{sec:recovery_rl}

\paragraph{Trajectory-level optimization.}
Starting from \(\pi_{\mathrm{sft}}\), we sample \(G\) trajectories per task and evaluate their final repository states with an executable
verifier. Let \(R_i\) denote the terminal reward of
\(\tau_i\). Following GRPO, we compute
\begin{equation}
    \widehat A_i
    =
    \frac{
        R_i-\mu_R
    }{
        \sigma_R+\epsilon
    },
    \qquad
    \mu_R
    =
    \frac{1}{G}
    \sum_{j=1}^{G}R_j,
    \label{eq:trajectory_advantage}
\end{equation}
where \(\sigma_R\) is the within-group reward standard deviation and
\(\epsilon\) is a numerical stabilizer.

The trajectory-level advantage evaluates overall task completion, but
cannot identify which intermediate recovery decisions contributed to the
outcome. Broadcasting it uniformly may reward an ineffective critique in
a successful trajectory or penalize a useful local correction in an
ultimately unsuccessful trajectory. We therefore construct one binary
process signal for each recovery stage.

\paragraph{Trigger signal.}
Let \(p_k\) be the position of the \(k\)-th self-critique and
\(\widehat b_k\) its chronologically matched teacher boundary. We define
\begin{equation}
\begin{aligned}
    h_k^{\mathrm{trig}}
    =
    \mathbb{1}
    \left[
        |p_k-\widehat b_k|
        \leq w
    \right],
\end{aligned}
\end{equation}
where \(w\) is a tolerance window measured in interaction steps. A
positive signal requires reflection to occur sufficiently close to the
earliest point at which the deviation can be reliably diagnosed.
Reflections without a valid boundary match receive a zero signal, while
unmatched boundaries have no local trigger span and remain governed by
the trajectory-level advantage. Appendix~\ref{app:matching} specifies the matching protocol.

\paragraph{Diagnosis signal.}
We evaluate a generated critique \(c_k\) against the reference
\(c_k^{\mathrm{ref}}\) in terms of error type, supporting evidence,
invalidated assumption, and recovery direction:
\begin{equation}
\begin{aligned}
    s_k^{\mathrm{crit}}
    =
    &\,
    \lambda_{\mathrm{type}}
    s_k^{\mathrm{type}}
    +
    \lambda_{\mathrm{evi}}
    s_k^{\mathrm{evi}}
    \\
    &+
    \lambda_{\mathrm{assump}}
    s_k^{\mathrm{assump}}
    +
    \lambda_{\mathrm{dir}}
    s_k^{\mathrm{dir}} .
\end{aligned}
\label{eq:critic_score}
\end{equation}
The component scores lie in \([0,1]\), and their non-negative weights sum
to one. Structured fields are compared directly, while open-ended
semantic fields are evaluated by a frozen consistency model. We do not
require lexical agreement with the reference, but all cited evidence must
be observable in the current prefix. The diagnosis signal is
\begin{equation}
\begin{aligned}
    h_k^{\mathrm{crit}}
    =
    \mathbb{1}
    \big[
        &s_k^{\mathrm{crit}}\geq\eta
        \\
        &{}\land
        \operatorname{Grounded}
        (c_k,h_{p_k})
    \big],
\end{aligned}
\label{eq:critic_signal}
\end{equation}
where \(h_{p_k}\) is the prefix available when \(c_k\) is generated and
\(\eta\) is the diagnostic-consistency threshold. Unsupported or
future-dependent diagnoses receive a zero signal.

\paragraph{Corrective-action signal.}
A correct critique is useful only if it changes subsequent behavior. To
measure this effect, we execute a training-time snapshot test after every
code-writing operation. Snapshot results are not exposed to the agent and
do not affect the original trajectory.

For each critique, let \(\mathcal P_k^{-}\) and
\(\mathcal P_k^{+}\) denote the passing-test sets of the nearest snapshots
before and after its corrective-action span. We define
\begin{equation}
\begin{aligned}
    F2P_k
    &=
    \mathcal P_k^{+}
    \setminus
    \mathcal P_k^{-},
    \\
    P2F_k
    &=
    \mathcal P_k^{-}
    \setminus
    \mathcal P_k^{+},
\end{aligned}
\label{eq:test_transition}
\end{equation}
where \(F2P_k\) represents repaired tests and \(P2F_k\) represents newly
introduced regressions.

Let \(A_k^{\mathrm{ca}}\) denote the action sequence covered by
\(T_k^{\mathrm{ca}}\). Because a test transition may be unrelated to the
critique, a frozen attribution judge examines the critique, its
subsequent actions, and the observed transition:
\begin{equation}
\begin{aligned}
    h_k^{\mathrm{ca}}
    =
    \mathbb{1}
    \big[
        \operatorname{Attr}
        (&c_k,A_k^{\mathrm{ca}},
        \\
         &F2P_k,P2F_k)
        =
        \texttt{effective}
    \big].
\end{aligned}
\label{eq:corrective_action_signal}
\end{equation}
A positive signal requires the actions to follow the diagnosed recovery
direction, produce attributable improvement, and avoid attributable
regression. If no post-critique write or attributable improvement is
observed, the signal is zero. The judge produces only this discrete,
test-grounded attribution label rather than an unconstrained scalar
reward.

\paragraph{Recovery-stage credit assignment.}
The three process signals do not replace the terminal reward. Instead,
they determine whether each recovery stage should inherit the positive
or negative component of the trajectory-level advantage. We define
\([\widehat A_i]^+=\max(\widehat A_i,0)\) and
\([\widehat A_i]^-=\min(\widehat A_i,0)\). For recovery stage
\(m\in\{\mathrm{trig},\mathrm{crit},\mathrm{ca}\}\), the shaped advantage
is
\begin{equation}
\widetilde A_{i,t}=h_k^m[\widehat A_i]^+
 +(1-h_k^m)[\widehat A_i]^-,\quad t\in T_k^m.
\label{eq:stage_advantage}
\end{equation}

A valid recovery decision can inherit positive credit from a successful
trajectory without inheriting negative credit from an unsuccessful one.
Conversely, an invalid decision can inherit negative credit from a failed
trajectory but cannot benefit merely from appearing in a successful one.
Tokens outside recovery spans retain $\widetilde A_{i,t}=\widehat A_i$.

We optimize the clipped token-level surrogate
\begin{equation}
\ell_{i,t}(\theta)=\min\!\left(
\rho_{i,t}\widetilde A_{i,t},\,
\bar\rho_{i,t}\widetilde A_{i,t}\right).
\label{eq:clipped_surrogate}
\end{equation}
where
\(\rho_{i,t}
=\pi_{\theta}(a_{i,t}\mid h_{i,t})/
\pi_{\theta_{\mathrm{old}}}(a_{i,t}\mid h_{i,t})\)
and
\(\bar{\rho}_{i,t}
=\operatorname{clip}
(\rho_{i,t},1-\varepsilon,1+\varepsilon)\).
Here, \(t\) indexes policy-generated tokens and \(|\tau_i|\) denotes the
number of such tokens in trajectory \(\tau_i\). The final objective is
\begin{equation}
\begin{aligned}
\mathcal J_{\mathrm{RIPPLE}}(\theta)
&=\mathbb E\!\left[\frac{1}{G}\sum_{i=1}^{G}
\frac{1}{|\tau_i|}\sum_t\ell_{i,t}(\theta)\right]\\
&\quad-\beta\mathcal D_{\mathrm{KL}}.
\end{aligned}
\label{eq:ripple_objective}
\end{equation}
Here, \(\mathcal D_{\mathrm{KL}}\) regularizes the current policy toward a
frozen reference.

The objective changes credit assignment without adding scalar process reward. If every rollout in a group has the same terminal reward, $\widehat A_i=0$ and sign gating cannot create a policy-gradient learning signal. Snapshot execution and frozen evaluators are training costs; deployment uses the self-recovering policy alone.

\section{Experiments}\label{sec:exp}
We ask whether \textsc{RIPPLE} improves end-to-end performance without an online critic (RQ1), whether recovery credit requires both locality and stage alignment (RQ2), and whether learned critiques functionally guide recovery (RQ3).

\subsection{Experimental Setup}
\paragraph{Actor and benchmarks.}
We use Qwen3.5-27B in a common SWE-agent-style edit--execute scaffold. Methods share prompts, tools, decoding, and a cap of 150 tool turns. Self-critique tokens count toward the actor-token budget. Post-training and evaluation repositories are disjoint. We report single-trajectory Resolved@1 on SWE-bench Verified \citep{jimenez2024swebench}, Pro \citep{deng2025swebenchpro}, and Multilingual \citep{khandpur2025swebenchmultilingual}. SWE-Compass uses its official task-aligned macro average \citep{xu2025swecompass}; we retain its separate metric. F2P/P2F analyses use the executable repair-like subset.

\paragraph{Comparison design.}
Test-time compute controls retain the tool-call cap while increasing the token budget. Controlled training variants start from the same Recovery-SFT checkpoint, $\pi_{\mathrm{sft}}$, and share tasks, verifier, optimizer, and budget. They adapt outcome-only GRPO, pivotal-aware credit \citep{guo2026pivoarl}, MURPHY-style retrospective credit \citep{ekbote2025murphy}, and ICRL-style internalization \citep{lin2026icrl}; each is a controlled mechanism adaptation rather than a native reproduction. Online-critic configurations are deployment references with additional model calls and are not compute matched. Trained methods average three training seeds; fixed methods average three matched decoding runs.

\paragraph{Fixed recovery evaluation.}
Before training any comparison method, we freeze the base policy and mine a Recovery Challenge Set of 800 recoverable and 400 non-deviating prefixes, each with a repository snapshot and remaining budget. All policies receive identical states. A separate evaluator from another model family labels boundaries and root causes; real execution supplies test transitions. A human-audited subset yields Cohen's $\kappa$ of 0.81, 0.76, and 0.73 for boundary, diagnosis, and action attribution, respectively.

We measure trigger precision/recall/F1 within $w=4$ tool turns of an annotated boundary; Diagnosis Accuracy; Action Alignment within $K=8$ turns; Functional Recovery, requiring $|\mathrm{F2P}|>0$ and $|\mathrm{F2P}|-|\mathrm{P2F}|>0$; and Full-Loop Recovery, requiring all four stages to succeed. Grounded Evidence measures whether cited evidence occurs in the prefix and is auxiliary: real evidence can still support an incorrect diagnosis. Appendix~\ref{app:evaluation-protocol} gives segmentation, matching, metrics, and the repository-clustered statistical protocol.

\subsection{RQ1: End-to-End Effectiveness}
\begin{table*}[t]
\centering
\small
\setlength{\tabcolsep}{3.0pt}
\begin{tabular}{lccccccc}
\toprule
& \multicolumn{3}{c}{Deployment Cost} & \multicolumn{4}{c}{Benchmark Performance} \\
\cmidrule(lr){2-4}\cmidrule(lr){5-8}
\raisebox{4pt}[0pt][0pt]{Method} & \raisebox{4pt}[0pt][0pt]{Ext.\ calls} & \raisebox{4pt}[0pt][0pt]{Act.\ tok.} & \raisebox{4pt}[0pt][0pt]{p95 lat.} & \shortstack{SWE-bench\\Verified} & \shortstack{SWE-bench\\Pro} & \shortstack{SWE-bench\\Multilingual} & \shortstack{SWE-\\Compass} \\
\midrule
Qwen3.5-27B Base & 0 & $1.00\times$ & $1.00\times$ & $74.6$ & $50.8$ & $68.8$ & $24.3$ \\
\midrule
\multicolumn{8}{l}{\emph{Test-time compute controls (identical tool-call budget, inflated token budget)}} \\
Base, longer single trajectory & 0 & $1.16\times$ & $1.21\times$ & $74.7$ & $51.7$ & $69.3$ & $25.0$ \\
Base, two restarts (split budget) & 0 & $1.17\times$ & $1.24\times$ & $75.0$ & $53.2$ & $70.1$ & $26.3$ \\
Prompted periodic reflection & 0 & $1.18\times$ & $1.24\times$ & $74.5$ & $52.6$ & $69.7$ & $25.6$ \\
\midrule
\multicolumn{8}{l}{\emph{Controlled process-credit variants (no auxiliary model served at deployment)}} \\
Recovery-SFT ($\pi_{\mathrm{sft}}$; RL init) & 0 & $1.19\times$ & $1.26\times$ & $75.1$ & $53.8$ & $70.7$ & $26.9$ \\
Outcome-only GRPO & 0 & $1.06\times$ & $1.09\times$ & $75.0$ & $54.5$ & $71.1$ & $27.5$ \\
Pivotal-aware credit & 0 & $1.09\times$ & $1.13\times$ & $75.2$ & $55.3$ & $71.8$ & $28.6$ \\
MURPHY-style retrospective & 0 & $1.09\times$ & $1.13\times$ & $75.2$ & $55.7$ & $72.3$ & $28.9$ \\
ICRL-style internalization & 0 & $1.15\times$ & $1.18\times$ & $75.4$ & $56.0$ & $72.7$ & $29.4$ \\
\textbf{\textsc{RIPPLE}} & \textbf{0} & $1.17\times$ & $1.23\times$ & $\mathbf{75.9}$ & $\mathbf{57.3}$ & $\mathbf{73.6}$ & $\mathbf{30.7}$ \\
\midrule
\multicolumn{8}{l}{\emph{Online-critic references (an auxiliary model is served; not compute matched)}} \\
Online same teacher & 4.8/task & $1.00\times$ & $1.57\times$ & $76.0$ & $57.9$ & $73.8$ & $31.0$ \\
Online strong critic & 10.9/task & $1.00\times$ & $2.61\times$ & $76.4$ & $58.4$ & $74.4$ & $31.6$ \\
\bottomrule
\end{tabular}
\caption{ End-to-end results and deployment cost. The first three benchmarks use Resolved@1; SWE-Compass uses its official task-aligned macro average. Actor tokens and p95 latency are realized quantities normalized to the base policy. Values are averages over three training seeds for trained methods and three matched decoding runs for fixed methods.}
\label{tab:main}
\end{table*}
Table~\ref{tab:main} reports gains over the base policy of 1.3 points on Verified, 6.5 on Pro, 4.8 on Multilingual, and 6.4 on SWE-Compass. The arithmetic mean improvement is 4.8 points after rounding. The repository-clustered 95\% interval crosses zero on Verified but excludes zero on the other three benchmarks (Appendix~\ref{app:rq1}). The smaller Verified gain is consistent with its lower prevalence of recoverable deviations.

Gains track opportunities for recovery. Recoverable deviations occur in 17.8\% of frozen base-policy trajectories on Verified, compared with 43.6\% on Pro, 39.8\% on Multilingual, and 41.2\% on the executable SWE-Compass subset. The reported repository-cluster association between deviation density and gain is $r=0.68$. On Pro, gains increase from 1.8 points in the Easy difficulty quartile to 11.9 in the Extreme quartile. Difficulty uses gold-patch size, files touched, module span, and test cost rather than evaluated-policy trajectories. Task, language, and distribution-shift slices appear in Appendix~\ref{app:rq1}. The largest task-slice gains occur in Bug Fixing and Feature Implementation, while C/C++/Go/Rust exhibit the largest language-group gain. These patterns are consistent with recovery benefiting from diagnostic feedback, although the grouped comparisons do not isolate the effect of feedback type.

On Pro, \textsc{RIPPLE} repairs 8.8\% of all instances that the base policy fails while damaging 2.3\% that it solves, yielding the 6.5-point net gain. Periodic reflection repairs 6.6\% and damages 4.8\% (Appendix~\ref{app:repair-damage}).

Additional computation alone does not reproduce the gain. At approximately $1.17\times$ realized actor tokens, longer execution and split-budget restarts reach 51.7 and 53.2 on Pro, versus 57.3 for \textsc{RIPPLE}. \textsc{RIPPLE} uses $1.17\times$ actor tokens and $1.23\times$ p95 latency relative to the base policy. It remains within 0.6 points of the online same-teacher reference on each benchmark, while eliminating auxiliary deployment calls. Internalization therefore removes recurring auxiliary-model calls while retaining the actor-side cost of self-critique.

\subsection{RQ2: Locality and Stage Alignment}
\begin{figure}[t]
\centering
\includegraphics[width=\columnwidth]{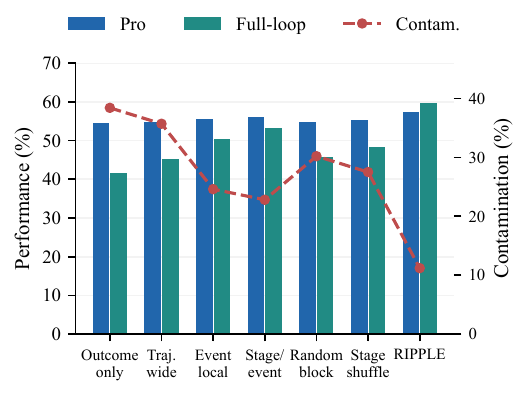}
\caption{\textbf{Locality and stage alignment contribute separately.} Seven credit-assignment variants use identical rollouts and process labels. Bars report Pro Resolved@1 and Full-Loop Recovery; the dashed line reports positive-credit contamination (lower is better). Event and Stage/event denote event-local generic and stage-aware event-wide credit; Random and Shuffle are sparsity controls.}
\label{fig:rq2}
\end{figure}
Equation~\ref{eq:stage_advantage} gates trajectory-advantage signs by local stage validity without adding scalar reward. Figure~\ref{fig:rq2} compares seven credit rules under the same rollouts and process labels. Restricting generic credit from the entire trajectory to a recovery event raises Full-Loop Recovery from 45.1\% to 50.4\% and Pro from 54.8 to 55.6. Distinguishing the stages but broadcasting their signals across the event reaches 53.1\% and 56.1. Applying each signal only to its corresponding stage reaches 59.6\% and 57.3. A recovery episode is therefore not credit-homogeneous: a timely trigger need not produce a correct diagnosis, and a correct diagnosis need not yield an effective edit.

Random-block gating updates the same number of tokens at randomly chosen positions. Stage-shuffled gating preserves the spans but permutes their signals. They reach only 45.8\% and 48.4\% Full-Loop Recovery (54.9 and 55.2 on Pro), suggesting that sparsity alone does not explain stage alignment. Positive-credit contamination decreases from 38.4\% for outcome-only GRPO to 24.6\% for event-local credit and 11.2\% for \textsc{RIPPLE}; Appendix~\ref{app:credit} records the complete ladder and clarifies the interpretation of this diagnostic.

Leave-one-stage-out results further localize the effect. Removing diagnosis credit reduces Diagnosis Accuracy from 81.5\% to 69.8\% and Action Alignment from 74.8\% to 64.3\%. Removing action credit leaves diagnosis near 80.9\% but lowers alignment to 56.8\%. Corrective-action credit has the largest end-to-end contribution: removing it lowers Pro from 57.3 to 54.0 and Full-Loop Recovery from 59.6\% to 43.5\% (Appendix~\ref{app:leave-one-out}).

\subsection{RQ3: Functional Recovery Mechanisms}
\begin{table}[t]
\centering
\small
\setlength{\tabcolsep}{2.0pt}
\begin{tabular}{lccccc}
\toprule
Method & Trig.\ F1 & Diag. & Align. & Func. & Full \\
\midrule
Periodic reflection & 0.43 & 55.8 & 48.1 & 42.5 & 31.4 \\
Recovery-SFT & 0.56 & 64.9 & 54.8 & 49.4 & 38.1 \\
Outcome-only & 0.60 & 66.9 & 58.6 & 53.0 & 41.6 \\
ICRL-style & 0.65 & 73.2 & 64.4 & 60.5 & 50.8 \\
\textbf{\textsc{RIPPLE}} & \textbf{0.79} & \textbf{81.5} & \textbf{74.8} & \textbf{70.2} & \textbf{59.6} \\
\bottomrule
\end{tabular}
\caption{Stage-wise recovery quality on the fixed Recovery Challenge Set, scored by an independent evaluator and real test transitions. Columns are trigger F1, Diagnosis Accuracy, Action Alignment, Functional Recovery, and Full-Loop Recovery; the last four are percentages.}
\label{tab:process}
\end{table}
\begin{figure}[t]
\centering
\includegraphics[width=\columnwidth]{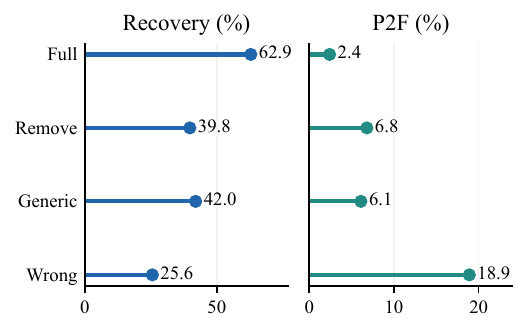}
\caption{\textbf{Matched-prefix critique intervention.} Holding policy, prefix, repository snapshot, remaining budget, and paired seeds fixed, the left panel reports $K$-step recovery and the right panel pass-to-fail regression for four critique conditions. The controlled contrasts show that a full critique outperforms both removal and a length- and style-matched generic critique, whereas a wrong-direction critique is worse than removal.}
\label{fig:rq3}
\end{figure}
\paragraph{Stage-wise behavior.}
Table~\ref{tab:process} evaluates local recovery on shared frozen states. \textsc{RIPPLE} leads every stage, reaching 74.8\% Action Alignment, 70.2\% Functional Recovery, and 59.6\% Full-Loop Recovery. Grounded Evidence is 84.0\%, compared with 81.5\% Diagnosis Accuracy, reinforcing the distinction between citing an observed failure and identifying its cause. On the 400 non-deviating prefixes, its false-trigger rate is 8.9\%, versus 71.5\% for periodic reflection. The fixed-state comparison evaluates abstention independently of each policy's on-policy deviation distribution. During training, outcome-only GRPO reduces the trigger rate from 0.180 to 0.110, below the reference deviation rate of 0.15, whereas \textsc{RIPPLE} converges to 0.142. This pattern suggests suppression of recovery opportunities rather than improved trigger calibration.

\paragraph{Matched-prefix intervention.}
Figure~\ref{fig:rq3} holds the policy checkpoint, prefix, repository snapshot, remaining budget, and paired decoding seed fixed while changing only the critique. Removing the critique lowers $K$-step recovery from 62.9\% to 39.8\% and raises P2F regression from 2.4\% to 6.8\%. A length- and style-matched generic critique reaches 42.0\% recovery and 6.1\% regression. A fluent but wrong-direction critique is worse than removal, at 25.6\% recovery and 18.9\% regression. These contrasts support a functional role for critique semantics beyond extra tokens or a pause in execution.

Extended interventions reach the same conclusion (Appendix~\ref{app:extended-intervention}). Direction alone retains 15.6 of the full critique's 23.1 recovery-point gain over removal, whereas evidence alone retains 4.9. Delaying the full critique by five turns loses 16.9 recovery points. The internal critique remains within 3.6 recovery points of the teacher reference. Snapshot tests are hidden from the policy, and these conclusions are conditional on the frozen states and intervention design.

\paragraph{Prefix supervision and recoverability.}
A prospective sequential prefix scan obtains boundary F1 of 0.84, compared with 0.86 for the offline-mined prefix pipeline and 0.90 for a full-trajectory oracle (Appendix~\ref{app:teacher-data}). Thus, prefix-only inputs retain much of the teacher's boundary-detection ability, although hindsight still improves the oracle. Randomly corrupting 10\% and 20\% of process labels lowers Pro by 0.5 and 1.2 points, respectively.

Recovery also depends on the available evidence and repository state. Among 248 prefixes with sufficient diagnostic evidence and a state judged reversible within the remaining budget, \textsc{RIPPLE} improves execution-verified resolution by 20.4 points over the base policy. Gains are 5.8 points when evidence is sufficient but the state is not reversible, 4.7 when evidence is insufficient but the state is reversible, and 1.7 when neither holds. Test, runtime, compiler, and implicit cross-file evidence show similar gains within the sufficient-and-reversible subset (Appendix~\ref{app:recoverability}). These results delineate the recoverability boundary: an effective handoff requires both a diagnosis supported by the observed prefix and a repository state that can still be repaired. The evidence-family decomposition further indicates that recovery is not restricted to explicit test failures; implicit cross-file inconsistencies can also support correction when both conditions hold.

\section{Conclusion}\label{sec:Conclusion}
We identify the Critique-to-Action Handoff Gap in long-horizon coding agents and propose \textsc{RIPPLE} to internalize prefix-only recovery guidance. Stage-specific credit distinguishes reflection triggering, diagnosis, and corrective action while retaining terminal task supervision. Local validity gates prevent ineffective recovery decisions from inheriting positive trajectory credit, while protecting useful corrections from negative credit attributable to unrelated failures. Across four software-engineering benchmarks, \textsc{RIPPLE} improves the base policy by an arithmetic mean of 4.8 score points without an online critic. Credit-assignment controls and matched-prefix interventions support the functional contribution of learned critiques. Gains concentrate where diagnostic evidence is sufficient and repository states remain reversible, motivating improved exploration in uniform-reward groups and direct estimation of local recovery effects across evolving repositories, programming languages, and deployment environments. Jointly learning when to reflect, what to diagnose, and how to act turns recovery into an internal policy capability for converting new execution evidence into effective corrections to the current repository state.

\section*{Limitations}\label{sec:Limitation}
\paragraph{Scope and supervision.}
We evaluate Qwen3.5-27B in one coding scaffold; transfer across model families and domains remains untested. Repository-disjoint post-training does not exclude pretraining exposure. Process labels are approximate, and offline candidate nomination uses complete trajectories despite prefix-only teacher inputs. Incomplete tests, flaky execution, and evaluator errors can affect attribution.

\paragraph{Cost and recovery boundaries.}
Auxiliary supervision and snapshot tests add training cost, for which we provide no complete amortized analysis. Deployment uses $1.17\times$ actor tokens and $1.23\times$ p95 latency relative to the base policy. Sign gating supplies no learning signal in uniform-reward groups and cannot overcome missing evidence or irreversible states within a fixed budget.

\paragraph{Evaluation limits.}
Fixed-prefix metrics and interventions are conditional on selected states and do not characterize each policy's on-policy failure distribution. The Verified gain's confidence interval crosses zero. The mean gain averages benchmark-specific score differences, rather than pooling issue-resolution rates; F2P/P2F analyses cover only executable repair-like tasks.

\bibliography{latex/custom}
\clearpage
\appendix
\raggedbottom
\setlength{\floatsep}{12pt}
\setlength{\textfloatsep}{14pt}
\setlength{\intextsep}{12pt}
\renewcommand{\topfraction}{0.95}
\renewcommand{\bottomfraction}{0.90}
\renewcommand{\textfraction}{0.05}
\makeatletter
\setlength{\@fptop}{0pt}
\setlength{\@fpsep}{12pt}
\setlength{\@fpbot}{0pt plus 1fil}
\makeatother

\providecommand{\RIPPLE}{\textsc{RIPPLE}}
\setcounter{equation}{0}
\setcounter{figure}{0}
\setcounter{table}{0}
\renewcommand{\theequation}{\thesection.\arabic{equation}}
\renewcommand{\thefigure}{\thesection.\arabic{figure}}
\renewcommand{\thetable}{\thesection.\arabic{table}}
\makeatletter
\@addtoreset{equation}{section}
\@addtoreset{figure}{section}
\@addtoreset{table}{section}
\makeatother

\lstdefinestyle{ripplejson}{
  basicstyle=\ttfamily\small,
  breaklines=true,
  columns=fullflexible,
  keepspaces=true,
  frame=single,
  framerule=0.3pt,
  xleftmargin=4pt,
  xrightmargin=4pt,
  aboveskip=4pt,
  belowskip=4pt,
  upquote=true,
  literate={"}{{\char34}}1,
  showstringspaces=false
}

\section{Recovery Supervision and Prefix-Only GRM Outputs}
\label{app:supervision}

This section supplies the construction details needed to audit the first
contribution: training-time recovery guidance is generated from trajectory
prefixes and is internalized into a single deployment policy. Complete offline
trajectories are used only to nominate bounded candidate regions. Every label
used by policy training is regenerated from a prefix record that excludes
future actions, the final patch, and the terminal verifier outcome.

\subsection{Candidate nomination and prefix regeneration}
\label{app:event-mining}

For a completed trajectory
\(\tau=(x,a_1,o_1,\ldots,a_T,o_T)\), offline mining proposes candidate regions
\(\mathcal W(\tau)=\{W_k\}_{k=1}^{M}\). Candidate anchors include newly failing
or changed tests, runtime exceptions, compiler or type-system diagnostics,
reproducible pass-to-fail transitions, repeated incompatible evidence, and
prefix-visible cross-file inconsistencies. Infrastructure failures, flaky
transitions, and deviations visible only from the future suffix are excluded.

For each \(t\in W_k\), the frozen generative reward model (GRM) receives only
\[
 h_t=(x,a_1,o_1,\ldots,a_{t-1},o_{t-1})
\]
and returns
\[
 g_\phi(h_t)=\bigl(q_t,\alpha_t,c_t^{\mathrm{ref}}\bigr),
\]
where \(q_t\in\{0,1\}\) indicates whether a recoverable deviation is observable, \(\alpha_t\in[0,1]\) is confidence, and \(c_t^{\mathrm{ref}}\) is a JSON record with causal analysis, issue categories, impact level, supporting evidence, invalidated assumption, and recovery direction. Recovery-window and verification-target fields provide auxiliary guidance. The event boundary is
\[
 \widehat b_k=\min\{t\in W_k:q_t=1,\ \alpha_t\geq\gamma\}.
\]
The GRM abstains when no prefix in the region supports a concrete recovery
direction. Prefix records are serialized independently, and an automated
field-level assertion rejects records containing suffix actions, final patches,
or terminal rewards.

\subsection{Illustrative GRM outputs}
\label{app:grm-examples}

The schema-valid examples below illustrate the information available to the
teacher; they are not evaluation instances and their confidence fields are not
used in any aggregate reported in the paper. The first example identifies a
configuration-precedence error after a failing test. The second abstains because
a transient tool failure does not yet falsify a repository-level assumption.

\begin{figure}[!htb]
\begin{lstlisting}[style=ripplejson]
{
 "recoverable": true,
 "analysis": "CLI timeout is parsed, then overwritten while RunnerConfig is built.",
 "issue_categories": [
  "configuration precedence",
  "state propagation"
 ],
 "impact_level": "high",
 "evidence": [
  "t38: timeout test expected 5, got 30",
  "t41: default read after CLI parsing"
 ],
 "invalidated_assumption": "Parsed --timeout already reaches RunnerConfig.",
 "recovery_direction": "Restore CLI-over-default precedence in the config adapter.",
 "confidence": 0.94,
 "recovery_window": "immediate",
 "verification_target": "timeout precedence test"
}
\end{lstlisting}
\caption{Illustrative actionable teacher output. All cited evidence is prefix-present; the recovery direction does not expose a hidden patch.}
\label{fig:grm-examples}
\end{figure}
\begin{figure}[!htb]
\begin{lstlisting}[style=ripplejson]
{
 "recoverable": false,
 "analysis": "Transient service timeout; no reproducible repository failure.",
 "issue_categories": [
  "infrastructure transient"
 ],
 "impact_level": "low",
 "evidence": [
  "t19: package-index request timed out"
 ],
 "invalidated_assumption": null,
 "recovery_direction": null,
 "confidence": 0.88,
 "recovery_window": "defer",
 "verification_target": "repeat package-index request"
}
\end{lstlisting}
\caption{Illustrative teacher abstention. A transient tool failure does not establish a repository-level deviation. Both JSON examples are excluded from quantitative evaluation.}
\label{fig:grm-abstention}
\end{figure}

\subsection{Recovery-trajectory construction and leakage audit}
\label{app:teacher-data}

Task trajectories supervise ordinary repository interaction. Recovery
trajectories insert the GRM critique at \(\widehat b_k\) and retain a
state-consistent corrective continuation. The initialization objective is
\[
 \mathcal L_{\mathrm{init}}
 =\mathcal L_{\mathrm{SFT}}(\mathcal D_{\mathrm{task}})
 +\lambda_{\mathrm{rec}}
  \mathcal L_{\mathrm{SFT}}(\mathcal D_{\mathrm{rec}}).
\]
Candidate filtering excludes infrastructure failures, flaky transitions, and deviations that become observable only in the future suffix. Aggregate benchmark scores are not used to infer training-data counts.

\begin{center}
\begin{minipage}{\columnwidth}
\centering
\small
\setlength{\tabcolsep}{4.2pt}
\begin{tabular}{lr}
\toprule
Teacher-evaluation regime & Boundary F1 \\
\midrule
Offline-mined prefix pipeline & 0.86 \\
Prospective sequential prefix scan & 0.84 \\
Full-trajectory oracle & 0.90 \\
\bottomrule
\end{tabular}
\captionof{table}{Prefix-only audit. A prospective scan recovers most boundaries
without an offline candidate hint. The full-trajectory oracle quantifies the
value of hindsight that is intentionally unavailable to the training teacher.}
\label{tab:prefix-audit}
\end{minipage}
\end{center}

\section{Recovery Episodes, Challenge Set, and Metrics}
\label{app:evaluation-protocol}

This section supports the mechanism claims by fixing how recovery events are
segmented, matched, and scored independently of the training-time GRM.

\begin{table}[!htb]
\centering\small
\begin{tabularx}{\columnwidth}{>{\raggedright\arraybackslash}p{1.75cm}>{\raggedright\arraybackslash}X}
\toprule
Metric & Definition \\
\midrule
Trigger P/R/F1 & Emit \texttt{<recover>} within $w=4$ tool turns of an annotated boundary. Non-deviating prefixes contribute false positives. \\
Diagnosis Accuracy & Root cause and invalidated assumption agree with the independent evaluator; cited evidence occurs in the prefix. \\
Action Alignment & Actions follow the diagnosed direction within $K=8$ turns. \\
Functional Recovery & Real execution gives $|\mathrm{F2P}|>0$ and $|\mathrm{F2P}|-|\mathrm{P2F}|>0$. \\
Full-Loop Recovery & Trigger, diagnosis, alignment, and functional recovery all succeed. \\
Positive-credit contamination & Fraction of locally invalid blocks receiving positive advantage; lower is better. \\
\bottomrule
\end{tabularx}
\caption{Process metrics. Trigger scores range from 0 to 1; all other metrics are percentages. Grounded Evidence is auxiliary because prefix-present evidence can still support an incorrect diagnosis.}
\label{tab:metrics}
\end{table}

\subsection{Episode segmentation and chronological matching}
\label{app:matching}

The \(k\)-th recovery episode is
\[
 e_k=\bigl(T_k^{\mathrm{trig}},T_k^{\mathrm{crit}},T_k^{\mathrm{ca}}\bigr).
\]
The policy emits an explicit \texttt{<recover>} delimiter. The trigger span
contains the decision to initiate recovery, the critique span ends at the next
tool-directed action, and the corrective-action span contains the following
repository-directed actions up to the next critique, terminal action, or the
fixed action horizon. Malformed or consecutive delimiters receive no positive
local signal.

Let \(P=(p_1<\cdots<p_J)\) be generated critique positions and
\(B=(\widehat b_1<\cdots<\widehat b_M)\) be teacher boundaries. We use an
order-preserving one-to-one matching within \(w\) turns that first maximizes
the number of pairs and then minimizes total temporal distance. An unmatched
critique receives \(h_k^{\mathrm{trig}}=0\); an unmatched teacher boundary does
not create an artificial token block.

Snapshot tests are run after code-writing actions during training but are
hidden from the policy. For the nearest valid snapshots before and after a
corrective span,
\[
 \mathrm{F2P}_k=P_k^+\setminus P_k^-,\qquad
 \mathrm{P2F}_k=P_k^-\setminus P_k^+.
\]
The frozen attribution judge emits only
\texttt{effective}/\texttt{ineffective}. A positive corrective-action label
requires directionally consistent actions, attributable F2P improvement, and
no attributable regression.

\subsection{Fixed Recovery Challenge Set}
\label{app:challenge-set}

Before any comparison method is trained, the base policy is frozen and used to
mine 800 recoverable prefixes and 400 non-deviating prefixes. Each item stores
the issue, action--observation prefix, repository snapshot, remaining budget,
and executable test state. Every policy receives the same frozen states, so the
metrics measure local recovery ability rather than each policy's endogenous
deviation distribution.

The training-time GRM is not reused for scoring. An evaluator from a different
model family labels boundaries and root causes; test transitions come from real
execution. A human-audited subset yields Cohen's \(\kappa=0.81\), \(0.76\), and
\(0.73\) for boundary, diagnosis, and action attribution, respectively.

\subsection{Statistical protocol}
\label{app:statistics}

End-to-end uncertainty is estimated with a repository-clustered paired
bootstrap; paired binary changes are additionally tested with McNemar's test.
The protocol corrects the four benchmark-level primary contrasts with Holm's method. Clustered confidence intervals are reported separately from multiplicity-adjusted hypothesis tests.
Intervention comparisons hold the prefix, repository snapshot, remaining
budget, policy checkpoint, and decoding seed fixed. All reported intervals are
computed over repository clusters rather than treating individual trajectories
from the same repository as independent.

\section{Training and Stage-Specific Credit Details}
\label{app:training-details}

All main comparisons use Qwen3.5-27B inside the same SWE-agent-style
edit--execute scaffold, with identical prompts, tools, decoding, a cap of 150
tool turns, and one actor-token budget that includes self-critique tokens.
Evaluation repositories are disjoint from post-training repositories. Trained
methods use three training seeds; fixed methods use three matched decoding
runs. The recovery teacher, consistency model, snapshot side channel, and
attribution judge are frozen during RL and removed at deployment.

Equation~\ref{eq:stage_advantage} applies separate binary validity labels to trigger, diagnosis, and corrective-action spans. A locally invalid stage cannot inherit positive trajectory advantage through the gate, and a locally valid stage is protected from negative credit due to an unrelated suffix. Ordinary action tokens retain the original advantage. This does not add a process reward; a group with identical terminal rewards still has zero policy-gradient advantage.

\begin{table}[!htb]
\centering
\small
\setlength{\tabcolsep}{3.8pt}
\begin{tabular}{>{\raggedright\arraybackslash}p{2.4cm}>{\raggedright\arraybackslash}p{4.7cm}}
\toprule
Setting & Value \\
\midrule
Actor / scaffold & Qwen3.5-27B / common edit--execute loop \\
Tool-turn cap & 150 \\
Trigger tolerance & \(w=4\) turns \\
Action horizon & \(K=8\) turns \\
Seeds & 3 \\
Initialization & Shared Recovery-SFT checkpoint \\
Shared controls & Tasks, verifier, optimizer, budget, and process labels \\
Training-only modules & Frozen teacher, consistency model, snapshot tests, and attribution judge \\
Online auxiliary modules & None \\
\bottomrule
\end{tabular}
\caption{Implementation settings relevant to comparison validity and deployment
cost. Optimizer-level hyperparameters are shared across the controlled
credit-assignment variants.}
\label{tab:implementation}
\end{table}

\section{Additional End-to-End Analyses}
\label{app:rq1}

\subsection{Benchmark gains and uncertainty}

\begin{center}
\begin{minipage}{\columnwidth}
\centering
\small
\setlength{\tabcolsep}{3.7pt}
\begin{tabularx}{\columnwidth}{>{\raggedright\arraybackslash}Xrrrr}
\toprule
Benchmark & Base & \RIPPLE & Gain & 95\% CI \\
\midrule
Verified & 74.6 & 75.9 & +1.3 & $[-0.2, 2.8]$ \\
Pro & 50.8 & 57.3 & +6.5 & $[3.9, 9.1]$ \\
Multilingual & 68.8 & 73.6 & +4.8 & $[2.1, 7.5]$ \\
SWE-Compass & 24.3 & 30.7 & +6.4 & $[3.2, 9.6]$ \\
\bottomrule
\end{tabularx}
\captionof{table}{End-to-end gains with repository-clustered 95\% confidence intervals.
The arithmetic mean gain is \(4.75\) points, reported as \(+4.8\) after rounding.
The Verified interval crosses zero; the other three exclude zero.}
\label{tab:benchmark-ci}
\end{minipage}
\end{center}

\subsection{Difficulty and recoverable-deviation prevalence}
\begin{figure}[!htb]
\centering
\includegraphics[width=\columnwidth]{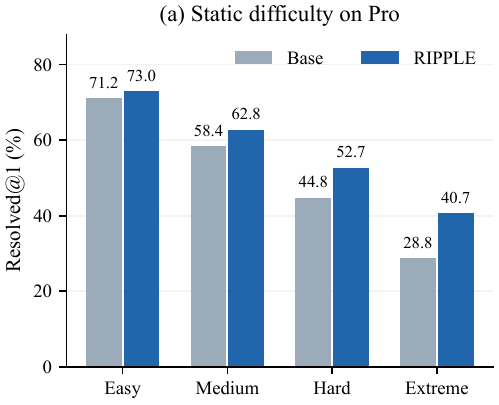}
\caption{Pro resolution by static-difficulty quartile. Gains increase from 1.8 points (Easy) to 11.9 (Extreme). Difficulty is defined without evaluated-policy trajectory statistics.}
\label{fig:rq1-additional}
\end{figure}

Figure~\ref{fig:rq1-additional} reports both analyses. Easy, Medium, Hard, and Extreme quartiles have Base/RIPPLE scores of 71.2/73.0, 58.4/62.8, 44.8/52.7, and 28.8/40.7. Their equal-weight averages reproduce Pro's 50.8/57.3 scores. Recoverable-deviation prevalence is 17.8\%, 43.6\%, 39.8\%, and 41.2\% on Verified, Pro, Multilingual, and the executable Compass subset, respectively.

\begin{figure}[!htb]
\centering
\includegraphics[width=\columnwidth]{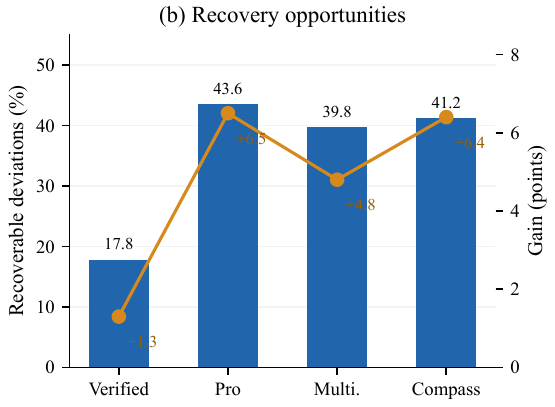}
\caption{Recoverable-deviation prevalence and benchmark gain. The reported $r=0.68$ association is computed over repository clusters, not these four aggregate points.}
\label{fig:opportunity}
\end{figure}

\subsection{Task, language, and distribution-shift slices}
Table~\ref{tab:slices} gives grouped task, language, and metadata-shift slices. The reported weights and rounded values reproduce the benchmark aggregates up to rounding. The slices locate performance differences without assigning a causal role to language or feedback type.

\begin{table}[!htb]
\centering\small
\begin{tabularx}{\columnwidth}{>{\raggedright\arraybackslash}Xrrr}
\toprule
Slice (weight) & Base & \RIPPLE & Gain \\
\midrule
\multicolumn{4}{l}{\emph{SWE-Compass}}\\
Bug fixing (1/8) & 23.4 & 32.2 & +8.8 \\
Feature implementation (1/8) & 25.8 & 33.4 & +7.6 \\
Other task types (6/8) & 24.2 & 30.0 & +5.8 \\
\midrule
\multicolumn{4}{l}{\emph{Multilingual}}\\
C/C++/Go/Rust (4/9) & 61.4 & 69.4 & +8.0 \\
Other languages (5/9) & 74.7 & 77.0 & +2.3 \\
\midrule
\multicolumn{4}{l}{\emph{Pro metadata shift}}\\
Lower half (1/2) & 52.6 & 58.3 & +5.7 \\
Higher half (1/2) & 49.0 & 56.3 & +7.3 \\
\bottomrule
\end{tabularx}
\caption{Task, language, and distribution-shift slices. Weighted rows reproduce their benchmark aggregates up to rounding. Pro's metadata-shift score is fixed before inference and split at its median.}
\label{tab:slices}
\end{table}

\subsection{Repair-versus-damage decomposition}
\label{app:repair-damage}

For paired runs on the same Pro instance,
\(\operatorname{Repair}=\Pr(Y_m=1,Y_B=0)\) and
\(\operatorname{Damage}=\Pr(Y_m=0,Y_B=1)\). Both use all Pro instances as the
denominator, so Repair minus Damage equals the absolute gain.

\begin{table}[!htb]
\centering
\small
\setlength{\tabcolsep}{3.7pt}
\begin{tabular}{lrrr}
\toprule
Method & Repair & Damage & Net \\
\midrule
Longer single trajectory & 4.4 & 3.5 & +0.9 \\
Two restarts & 6.0 & 3.6 & +2.4 \\
Prompted periodic reflection & 6.6 & 4.8 & +1.8 \\
Recovery-SFT & 6.2 & 3.2 & +3.0 \\
Outcome-only GRPO & 6.6 & 2.9 & +3.7 \\
Pivotal-aware credit & 7.2 & 2.7 & +4.5 \\
MURPHY-style retrospective & 7.5 & 2.6 & +4.9 \\
ICRL-style internalization & 7.8 & 2.6 & +5.2 \\
\textbf{\RIPPLE} & \textbf{8.8} & \textbf{2.3} & \textbf{+6.5} \\
\bottomrule
\end{tabular}
\caption{Paired repair-versus-damage decomposition on Pro. Every row uses all
Pro instances as the denominator, so Repair minus Damage exactly equals the
Resolved@1 gain over the base policy. \RIPPLE{} achieves both the highest repair
rate and the lowest damage rate among the trained comparisons.}
\label{tab:repair-damage}
\end{table}

\section{Additional Credit-Assignment Analyses}
\label{app:credit}

\subsection{Complete locality and alignment ladder}

\begin{center}
\begin{minipage}{\columnwidth}
\centering
\small
\setlength{\tabcolsep}{3.2pt}
\begin{tabular}{lrrr}
\toprule
Credit assignment & Pro & Full-loop & Contam. \\
\midrule
Outcome-only & 54.5 & 41.6 & 38.4 \\
Trajectory-wide generic & 54.8 & 45.1 & 35.7 \\
Recovery-event-local generic & 55.6 & 50.4 & 24.6 \\
Stage-aware, event-wide & 56.1 & 53.1 & 22.8 \\
Random-block control & 54.9 & 45.8 & 30.2 \\
Stage-shuffled control & 55.2 & 48.4 & 27.5 \\
Stage-aligned (\RIPPLE) & 57.3 & 59.6 & 11.2 \\
\bottomrule
\end{tabular}
\captionof{table}{Values underlying the main-paper credit-assignment figure. All
variants use identical tasks, rollouts, process labels, and optimization
budgets. Contamination is lower-is-better.}
\label{tab:credit-ladder}
\end{minipage}
\end{center}

Under Equation~\ref{eq:stage_advantage}, a block with training label $h_k^m=0$ cannot receive positive shaped advantage. A nonzero contamination score therefore requires a validity assessment that differs from the gate's labels or coverage. This diagnostic must be interpreted separately from the algebraic guarantee of sign gating.

\subsection{Leave-one-stage-out analysis}
\label{app:leave-one-out}
\begin{figure}[!htb]
\centering
\includegraphics[width=\columnwidth]{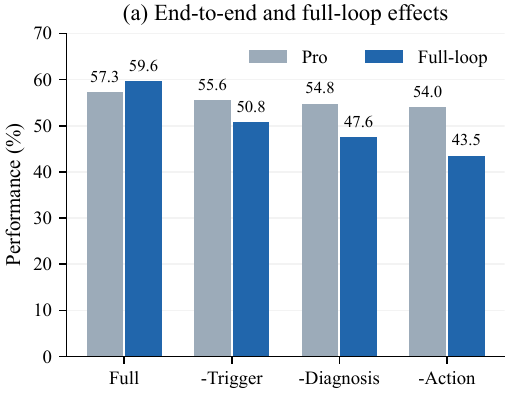}
\caption{Leave-one-stage-out end-to-end effects. Removing corrective-action credit produces the largest drop in Pro and Full-Loop Recovery.}
\label{fig:leave-one-out}
\end{figure}

Figure~\ref{fig:leave-one-out} removes one stage's aligned credit at a time. Removing trigger credit lowers trigger F1 from 0.79 to 0.67 while preserving diagnosis at 80.8\%; removing diagnosis lowers diagnosis to 69.8\% and alignment to 64.3\%; removing action credit preserves diagnosis at 80.9\% but lowers alignment to 56.8\%. The corresponding Pro scores are 55.6, 54.8, and 54.0, versus 57.3 for the full method. Full-loop scores are 50.8\%, 47.6\%, and 43.5\%, versus 59.6\%.

\begin{figure}[!htb]
\centering
\includegraphics[width=\columnwidth]{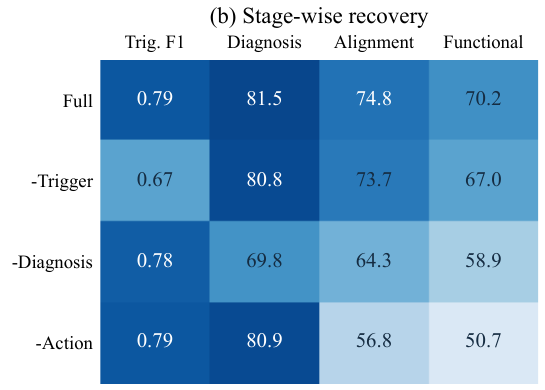}
\caption{Stage-wise ablation. Trigger credit primarily affects initiation; diagnosis credit affects diagnosis and downstream action; removing action credit leaves upstream stages largely unchanged. Trigger F1 is shown on a 0--1 scale; other cells are percentages.}
\label{fig:stage-matrix}
\end{figure}

\section{Mechanism Audits and Recoverability Boundary}
\label{app:mechanism}

\subsection{Extended matched-prefix intervention}
\label{app:extended-intervention}

The intervention fixes the policy checkpoint, trajectory prefix, repository snapshot, remaining tool-turn budget, and paired decoding seed. Only the critique inserted at the matched boundary changes. Recovery uses the same $K=8$-turn horizon as Action Alignment; P2F measures newly failing tests at the post-intervention snapshot.

The primary conditions separate semantic content from additional computation. \emph{Full critique} supplies the policy's diagnosis and recovery direction; \emph{remove critique} deletes the span without adding tokens; \emph{generic critique} matches its length and style but omits a repository-specific cause and correction; \emph{wrong-direction} supplies a fluent, prefix-compatible diagnosis with a correction that conflicts with the annotated root cause.

\begin{figure}[!htb]
\centering
\includegraphics[width=\columnwidth]{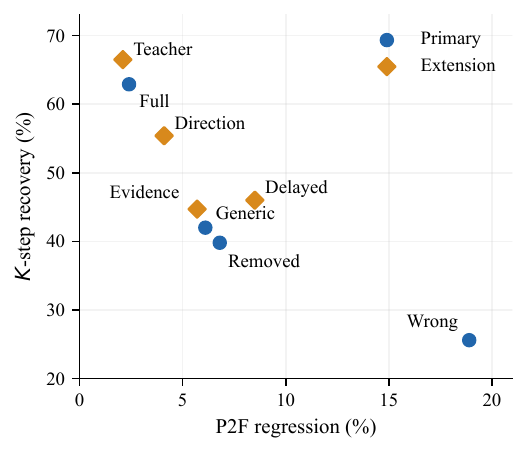}
\caption{Matched-prefix recovery--regression frontier. Better outcomes move toward the upper-left. Direction alone retains much of the benefit; delaying the critique forfeits a substantial share.}
\label{fig:extended-intervention}
\end{figure}

Four extensions test content, timing, and internalization. \emph{Direction only} removes the evidence narrative; \emph{evidence only} omits the corrective direction; \emph{delayed five turns} inserts the same full critique after additional actions; \emph{teacher reference} substitutes the prefix-only teacher's output. Figure~\ref{fig:extended-intervention} and Table~\ref{tab:extended-intervention} report all conditions.

Snapshot tests run after the rollout and remain hidden from the policy. The same test identifiers define F2P and P2F in every condition, and intervention prompts are fixed before observing outcomes. The comparison therefore holds task selection, state, and verifier access constant, but remains conditional on the selected prefixes.

Relative to removal, the full critique adds 23.1 recovery points and reduces P2F by 4.4 points. A generic critique adds only 2.2 recovery points, while a wrong direction loses 14.2 and adds 12.1 P2F points. Direction alone retains 15.6 of the full critique's 23.1-point benefit; evidence alone retains 4.9. Delay costs 16.9 recovery points and adds 6.1 P2F points. The internal critique is 3.6 recovery points below the teacher reference, with a 0.3-point P2F gap. These contrasts support transferring a timely corrective direction into subsequent actions. Process labels remain approximate and the conclusions are conditional on the fixed prefixes; the main Limitations section discusses test incompleteness, evaluator error, and broader scope.

\begin{table}[!htb]
\centering\small
\begin{tabularx}{\columnwidth}{>{\raggedright\arraybackslash}Xrr}
\toprule
Condition & Recovery & P2F \\
\midrule
Full critique & 62.9 & 2.4 \\
Remove critique & 39.8 & 6.8 \\
Generic critique & 42.0 & 6.1 \\
Wrong-direction critique & 25.6 & 18.9 \\
Direction only & 55.4 & 4.1 \\
Evidence only & 44.7 & 5.7 \\
Delayed five turns & 46.0 & 8.5 \\
Teacher reference & 66.5 & 2.1 \\
\bottomrule
\end{tabularx}
\caption{Extended matched-prefix interventions (\%). All conditions share the policy, frozen prefix, repository snapshot, remaining budget, and paired seed. The internal critique is within 3.6 recovery points and 0.3 P2F points of the teacher reference.}
\label{tab:extended-intervention}
\end{table}

\subsection{Label noise and selective triggering}

\begin{center}
\begin{minipage}{\columnwidth}
\centering
\small
\setlength{\tabcolsep}{4.5pt}
\begin{tabular}{lrr}
\toprule
Corrupted process labels & Pro & Drop \\
\midrule
0\% & 57.3 & 0.0 \\
10\% & 56.8 & 0.5 \\
20\% & 56.1 & 1.2 \\
\bottomrule
\end{tabular}
\captionof{table}{Label-noise audit. Randomly flipping 10--20\% of the process labels
degrades Pro smoothly and costs fewer than 1.5 points, indicating that the
stage-local objective does not depend on perfectly clean labels.}
\label{tab:label-noise}
\end{minipage}
\end{center}

On 400 non-deviating prefixes, \RIPPLE's false-trigger rate is 8.9\%, versus
71.5\% for prompted periodic reflection. This fixed-state contrast is more
diagnostic than raw critique frequency: a useful recovery policy must abstain
when no deviation is present.

\newpage
\subsection{Evidence sufficiency and state reversibility}
\label{app:recoverability}

The independent evaluator labels whether the prefix contains enough evidence
to identify a concrete root cause and whether the current repository state can
still be repaired within the remaining budget. The main paper reports that
when both conditions hold (\(n=248\)), \RIPPLE{} gains 20.4 points over the base
policy; when either fails, the gain is single-digit.

\begin{center}
\begin{minipage}{\columnwidth}
\centering
\small
\setlength{\tabcolsep}{3.5pt}
\begin{tabular}{ccrrrr}
\toprule
Evidence & Reversible & \(n\) & Base & \RIPPLE & Gain \\
\midrule
Yes & Yes & 248 & 42.7 & 63.1 & +20.4 \\
Yes & No & 174 & 31.6 & 37.4 & +5.8 \\
No & Yes & 206 & 27.2 & 31.9 & +4.7 \\
No & No & 172 & 12.8 & 14.5 & +1.7 \\
\bottomrule
\end{tabular}
\captionof{table}{Recoverability boundary on the 800 recoverable-prefix items. The
20.4-point gain requires both sufficient diagnostic evidence and a repository state judged reversible within the remaining budget; the reported gain is single-digit in every other cell.}
\label{tab:recoverability}
\end{minipage}
\end{center}

\begin{center}
\begin{minipage}{\columnwidth}
\centering
\small
\setlength{\tabcolsep}{3.0pt}
\begin{tabularx}{\columnwidth}{>{\raggedright\arraybackslash}Xrrrr}
\toprule
Evidence family & \(n\) & Base & \RIPPLE & Gain \\
\midrule
Test output & 96 & 44.8 & 65.6 & +20.8 \\
Stack trace / runtime & 54 & 39.0 & 58.6 & +19.6 \\
Compiler / type system & 60 & 41.7 & 63.4 & +21.7 \\
Implicit cross-file inconsistency & 38 & 44.2 & 62.7 & +18.5 \\
\bottomrule
\end{tabularx}
\captionof{table}{Decomposition of the 248 evidence-sufficient, reversible prefixes.
The weighted average gain is 20.4 points up to rounding, consistent with
Table~\ref{tab:recoverability}; explicit and implicit evidence support similar
improvements.}
\label{tab:evidence-family}
\end{minipage}
\end{center}

\end{document}